\pdfoutput=1

\documentclass[11pt]{article}

\usepackage[]{acl}

\usepackage{times}
\usepackage{latexsym}

\usepackage[T1]{fontenc}

\usepackage[utf8]{inputenc}

\usepackage{microtype}

\usepackage{inconsolata}

\usepackage{titlesec}
\usepackage{float}
\usepackage{graphicx}
\usepackage{amsmath}
\usepackage{datetime}
\usepackage{setspace}
\usepackage{todonotes}
\usepackage{fancyhdr}
\usepackage{url}
\usepackage{amsmath}

\title{A Truly Joint Neural Architecture for Segmentation and Parsing}

\author{Danit Yshaayahu Levi\qquad Reut Tsarfaty\\Bar Ilan University, Ramat Gan 5290002, Israel\\ \texttt{danity251@gmail.com}\qquad \texttt{reut.tsarfaty@biu.ac.il}}

\begin{document}
\maketitle
\begin{abstract}
Contemporary multilingual dependency parsers can parse a diverse set of languages, but for Morphologically Rich Languages (MRLs), performance is attested to be lower than other languages. The key challenge is that, due to high morphological complexity and ambiguity of the space-delimited input tokens, the linguistic units that act as nodes in the tree are not known in advance. Pre-neural dependency parsers for MRLs subscribed to the {\em joint morpho-syntactic hypothesis}, stating that morphological segmentation and syntactic parsing should be solved jointly, rather than as a pipeline where segmentation precedes parsing. However, neural state-of-the-art parsers to date use a strict pipeline. 
In this paper we introduce a joint neural architecture where a lattice-based representation preserving all morphological ambiguity of the input is provided to an arc-factored model, which then solves  the morphological segmentation and syntactic parsing tasks at once. Our experiments on Hebrew, a rich and highly ambiguous MRL, demonstrate state-of-the-art performance on parsing, tagging and segmentation of the Hebrew section of UD, using a single model. This proposed architecture is LLM-based and language agnostic, providing a solid foundation for MRLs to obtain further performance improvements and bridge the gap with other languages.
\end{abstract}

\section{Introduction} \label{sec:introduction}
Dependency parsing is the task of automatically analyzing the syntactic structure of a sentence and exposing the functional relationships between its words. 
In the past, dependency parsing was shown to be extremely useful for many language processing  tasks, as machine translation \cite{machine-translation}, question answering \cite{question-answering} and information extraction \cite{information-extraction}, to name a few. 
While nowadays many English NLP tasks are solved end-to-end using large language models (LLMs) and without accessing any symbolic structure, for low- and medium-resource languages, parsers are still indispensable, enabling a host of downstream applications.

Most neural state-of-the-art dependency parsers to date presuppose a pipeline architecture \cite{stanza, spacy, trankit} that includes several analysis stages ---  tokenization, word segmentation, part-of-speech (POS) tagging, morphological feature tagging, dependency parsing, and sometimes also named entity recognition --- and the linguistic features from each stage are provided as input to the tasks that follow it, and contribute to the overall efficacy. 
In morphologically-rich languages (MRLs), many raw space-delimited tokens consist of multiple units, each of which serves a distinct role in the overall syntactic representation \cite{tsarfaty-etal-2010-statistical, tsarfaty-etal-2020-spmrl}. Consequently, segmentation is essential for accurate MRL parsing. 
However, due to high morphological ambiguity, when segmentation is performed prior to (and independently of) the parsing phase, segmentation errors may propagate to undermine the syntactic predictions, and subsequently lead to an overall incorrect parse.

According to the joint hypothesis, that was heavily populated  in parsing studies for MRLs in the pre-neural era \cite{joint_hypothesis,cohen-smith-2007,goldberg-tsarfaty-2008,green-manning-2010,turkish}, morphological segmentation and syntactic predictions are mutually dependent, and hence, these two tasks should be solved together. 

Following these lines, \citet{yap} developed a joint morpho-syntactic transition-based parser that achieved state-of-the-art (SOTA) results on Hebrew parsing. This system employs a morphological lattice as input for a transition system, with both syntactic and morphological transitions, for picking the right arcs and segments in tandem. 
Another influential work is that of \citet{turkish}, that displays all potential segments that could be involved in any token's analysis, and allows an MST graph-based parser pick the highest scoring subset of arcs and nodes as the output dependency tree, showing parsing improvements for both Hebrew and Turkish.  
Similarly, \citet{krishna-etal-2020-keep,krishna-etal-2020-graph} contributed  a non-neural graph-based parser for Sanskrit which employ energy-based modeling to determine the optimal path on a graph which jointly represents a valid segmentation and syntactic analysis. However, this architecture is non-neural and Sanskrit specific, relying on labor-intensive hand-crafted feature engineering. 
In neural settings, and still for Sanskrit, \citet{sandhan2021little} introduced a pre-training approach which focuses on amalgamating word representations generated by encoders trained on auxiliary tasks, such as morphological and syntactic tags. Unlike the present work, this neural architecture does not make any segmentation decisions nor does it leverage the lattice structure for joint segmentation and parsing.

All in all, in the case of neural multilingual dependency parsers, the pipeline approach of segment-then-parse is fully maintained \cite{udpipe,udify,trankit}, and no language-agnostic architecture for morphological segmentation and syntactic parsing  is yet to be found. 

In this paper we revisit the {\em joint hypothesis} as a viable way to improve neural dependency parsing  for MRLs. 
The idea, in a nutshell, is as follows. We start off with an arc-factored model \cite{dozatmanning} that accepts a sequence of words as input, and generates  a dependency tree by picking the highest scoring arcs connecting all words. In our approach,
the arc-factored model  takes as input  a linearized lattice containing all possible morphological segments that may {\em potentially} act as nodes,  and learns to assign a head and label to each such node. 
During training, incorrect segments are  mapped to an auxiliary node, of which subtree is excluded from the final dependency tree. At inference time then, the model  maps relevant segments to the main branch and unused segments to the auxiliary branch,   building a complete dependency tree. In this process, morphological segmentation decisions get informed by the syntactic arcs, and vice versa.
We further extend the architecture with a multi-task learning (MTL) component to predict the  features of each node, e.g., {\em POS,} {\em gender, number} and {\em  person}.
 
Our experiments on the Hebrew Section of UD\footnote{The UD initiative  \url{https://universaldependencies.org/treebanks/he_htb/index.html}}
show that in cases where the input morphological analyses are complete, our model provides new and improved state of the art results for segmentation and parsing for Hebrew, in a single, jointly trained, model. In the more realistic case, where some of the word lattices may lack possible analyses (the case of out of vocabulary (OOV) tokens), the model still delivers competitive results for segmentation, tagging and parsing, outperforming the  state-of-the-art results of  the  de-facto standard pipelines, Stanza  and Trankit.

\section{Challenge and Research Objectives} \label{challenge}
The goal of dependency parsing is to automatically analyze the syntactic structure of a sentence by indicating the functional relationship between its words. The input is assumed to be a sequence of space-delimited tokens that represent words, and the output is a directed  tree where  each input word serves as a node, and each arc represents a relation between two such words. An arc can be labeled to indicate the relation type between the two words.

Deep neural networks have recently achieved unprecedented results in many areas of natural language processing, including the dependency parsing task. The architecture of \citet{dozatmanning} (that followed up on \citet{kiperwassergoldberg}) is currently accepted as the standard architecture for dependency parsing.
\citeauthor{dozatmanning} present a simple neural architecture where an arc factored model selects the best set of dependency arcs and labels. This approach is the foundation of several SOTA dependency parsers, including Stanza \cite{stanza} and Trankit \cite{trankit}, which have been trained and successfully used across different  languages. Crucially, these parsers  and others \cite{dozatmanning,stanza,trankit,udpipe,udify} all subscribe to a pipeline approach, where the input tokens are pre-segmented, and these segments  uniquely determine the nodes in the tree.

\begin{figure}[t]
    \centering
    \includegraphics[width=0.5\textwidth]{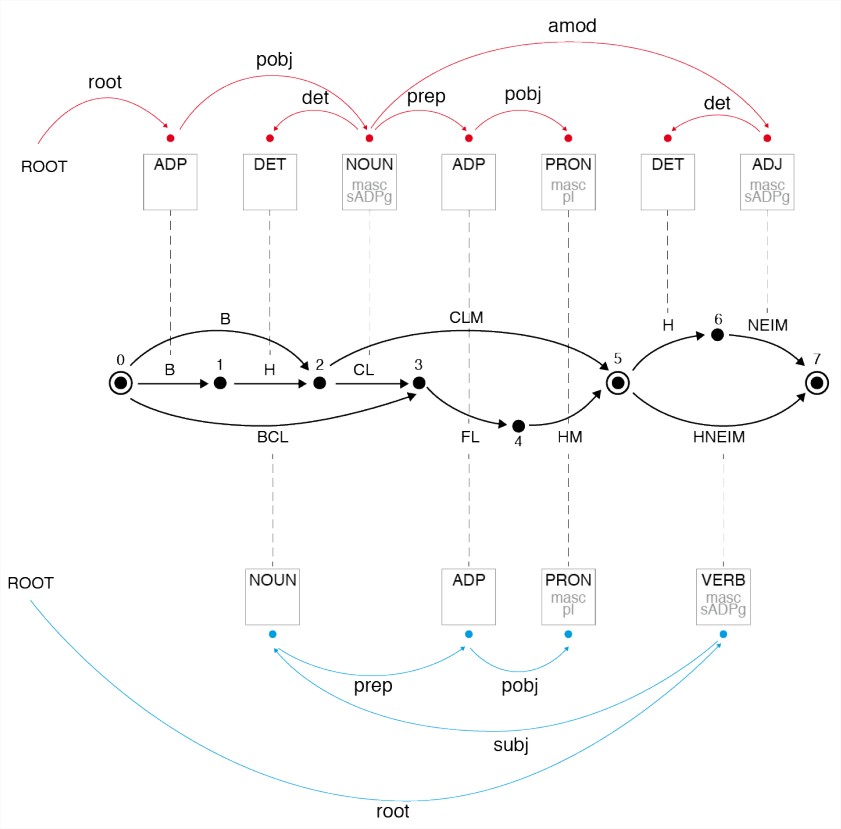}
    \caption{The morphological lattice for the Hebrew phrase \textit{bclm hneim} and two associated dependency trees depicting alternative segmentations (Origin: \citeauthor{yap} (\citeyear{yap})). The upper tree illustrates the syntactic structure corresponding to "In their pleasant shadow", while the lower tree corresponds to "Their onion was pleasant". This highlights the existence of multiple morphological decompositions and various potential dependency trees.
    }
    \label{fig:yap}
\end{figure}

This pipeline approach has been applied  across many language types, including morphologically rich languages (MRL). However, MRLs pose a significant challenge to such architectures.
In a pipeline architecture, where morphological segmentation is performed prior to parsing.
However, tokens in MRLs are rich and complex, and include multiple units that can act as individual nodes in the tree.  Hence their segmentation may be highly ambiguous, and the nodes of the tree are not known in advance.  When these segments are fixed prior to parsing, wrong segmentation seriously hinders parsing results. The main challenge is then to find the appropriate segmentation that is relevant to the particular syntactic context.
This challenge is illustrated at Figure~\ref{fig:yap}. 
here we consider the Hebrew phrase '\textit{bclm hneim}', which can be translated in various ways depending on the segmentation analysis applied: "In their pleasant shadow", "In the pleasant photographer," or "Their onion was pleasant". Figure \ref{fig:yap} provides a lattice representation of all morphological analyses of the phrase, where different segmentations give rise to substantially different syntactic trees.

Pre-neural models addressed this challenge by jointly modeling morphological segmentation and dependency parsing, and have shown that it yields superior results for both tasks. The pressing research question at hand is whether this hypothesis can also be validated within the context of neural parsing architectures. In other words, can neural parsing models benefit from a joint approach to segmentation and parsing, similar to what has been observed in non-neural models?

This paper addresses two primary objectives. Firstly, we aim to introduce a unified neural architecture that jointly solves segmentation, tagging and parsing, with the aim of  empirically validating the joint hypothesis within the realm of neural architectures. Secondly, we seek to attain state-of-the-art (SOTA) results for Hebrew, a language renowned for its formidable parsing challenges attributed to its substantial morphological ambiguity.
\section{The Proposal: A Model for Joint Morphological Segmentation and Syntactic Dependency Parsing}

\paragraph{Task Definition} 
Formally, our proposed model is defined as a structure prediction function $f: S \rightarrow D$, where $s \in S$ represents a sequence of raw input tokens, and $d \in D$ denotes a dependency tree with nodes corresponding to disambiguated units, which we refer to here as \textit{morphological segments}. 
Crucially, we retain morphological ambiguity, and deliver all possible analyses of \(s\) to the parser. 
Hence, we assume a Morphological Analyzer (MA), that given an input sentence $s=s_1,\dots,s_k$ yields a \textit{token-lattice} termed  $L_i=MA(s_i)$ for each token $s_i$. The complete lattice of the input sentence $L_s =MA(s)$ is defined as the concatenation of the token lattices $L_s = MA(s_1)\circ\dots\circ MA(s_k)$. Our structure prediction function  becomes   $f: L \rightarrow D$, with \(L_s \in L\) as the morphological lattice of \(s\in S\).

\paragraph{Input Linearization} \label{linerization}
Upon receiving an input lattice \(L_s\), we aim to linearize it in order to be able to encode it as an input vector for the neural architecture. As shall be seen shortly, the linearization is a critical phase for obtaining a neural encoding of the non-linear, morphologically ambiguous, input.
We illustrate the linearization process using the Hebrew sentence '\textit{bkrti bbit hlbn}' (lit: "I-visited in-the-house the-white", trans: "I visited the white house"). Initially, the $MA$ provides a list of all potential analyses for each token: bkrti: [('bkrti')], bbit: [('b', 'bit'), ('b', 'h', 'bit')], hlbn: [('h', 'lbn'), ('hlbn')]. Subsequently, each token is linearized independently: bkrti: ['bkrti'], bbit: ['b', 'bit', 'b', 'h', 'bit'], hlbn: ['h', 'lbn', 'hlbn']. 
Finally, all linearized analyses are concatenated according to the initial order: ['bkrti', 'b', 'bit', 'b', 'h', 'bit', 'h', 'lbn', 'hlbn'].

Formally, the \textit{linearize} function takes an input lattice $L_s$ and returns a sequence of $m$ analyses while maintaining the partial order of the tokens. Within the input lattice $L_s = MA(s_1)\circ\dots\circ MA(s_k)$, each $MA(s_j)$ encompasses a comprehensive set of potential analyses ---  segmentation options --- for the token $s_j$. Let $k_j$ be the number of analyses of the $j^{th}$ token; then, it holds that $\sum_{j=1}^{n} k_j = m$. Also, let $a_{j}^{i}$ be the $i^{th}$ analysis of the $j^{th}$ token. The linearization function works as follow:
\begin{gather*}
linearize(L_t) = \\
linearize(MA(t_1)\circ\dots\circ MA(t_k)) = \\
linearize(MA(t_1))\circ\dots\circ linearize(MA(t_k)) = \\
a_{1}^{1},\dots,a_{1}^{k_1}\circ 
a_{2}^{1},\dots,a_{2}^{k_2}\circ\dots\circ a_{n}^{k_n}    
\end{gather*}
The number of morphemes in an analysis $a_{i}^{j}$ is $r_{(i,j)}$, denoted as
\begin{gather*}
a_{i}^{j} = m_{(i,j)}^{1},\dots,m_{(i,j)}^{r_{(i,j)}-1},m_{(i,j)}^{r_{(i,j)}}
\end{gather*}
Consequently, the total number of morphemes in the linearized lattice is given by $\sum_{j=1}^{n}\sum_{i=1}^{k_{j}} r_{(i,j)}$. Thus, the linearized lattice can be expressed as a sequence of morphemes:
\begin{gather*}
linearize(L_t) = \\ m_{(1,1)}^{1},\dots,m_{(1,1)}^{r_{(1,1)}},\dots,m_{(n,k_n)}^{r_{(n,k_n)}}    
\end{gather*}

\paragraph{Joint Prediction} \label{para:joint_prediction}
We extend the simple and well-known neural arc-factored model to accept  a linearized Lattice as input, and choose a subset of arcs with the highest scores as the output dependency tree. Crucially, this selected set of arcs does not take all  segments as nodes. On the contrary, the arc selection essentially determines which segments from the lattice are included in the final tree, and is subject to certain constraints, as we detail shortly.



Let us define \(\mathcal{A}(L_s)\)  the set of all possible subsets of arcs in the linearized lattice. In our model we aim to select a highest scoring subset \(A\) as the DEP tree:
\[DEP = argmax_{A\in\mathcal{A}(L_s)} score (A)\]


To ensure that the nodes of the selected arcs in  \(A\) form a valid morpheme sequence, the  nodes that participate in the subset of arcs must adhere to   the following constraints:
\begin{enumerate}
\item Exactly one segmentation analysis should be chosen per token.
\item All morphemes in the chosen analysis should be included in the selection.
\item Arcs cannot connect morphemes from {\em different} analyses of {\em the same} token.
\end{enumerate}

When enforcing these constraints, the set of subsets  \(\mathcal{A}(L_s)\) is significantly smaller than a straightforward cartesian product over all possible segments. We thus define the set \(C\):
\begin{gather*}
C = constrained(\mathcal{A}(L_s)) = \{a_{1}^{1},\dots,a_{1}^{k_1}\}\times \\
\{a_{2}^{1},\dots,a_{2}^{k_2}\}\times\dots\times \{a_{n}^{1},\dots,a_{n}^{k_n}\}  
\end{gather*}
And the prediction function becomes
\[DEP = argmax_{A\in C} score (A)\]

Finally, note that in this model, \(A\) is not formally defined to form a tree. In order to ensure a tree structure, at inference time we employ an Maximum Spanning Tree algorithm (MST) on top of the constrained graph. 
\[DEP = argmax_{A\in C} MST\_score (A)\]
Note that the highest scoring tree uniquely defines the set of nodes that participate in it, so the MST in this proposed method also acts to substantiate the scoring function for morphological disambiguation (MD). We thus get: 
\[\langle MD,DEP\rangle = argmax_{A\in C} MST\_score (A)\]
\section{The Overall Architecture} \label{sec:architecture}

\begin{figure}[t]
    \centering
    \includegraphics[width=0.47\textwidth]{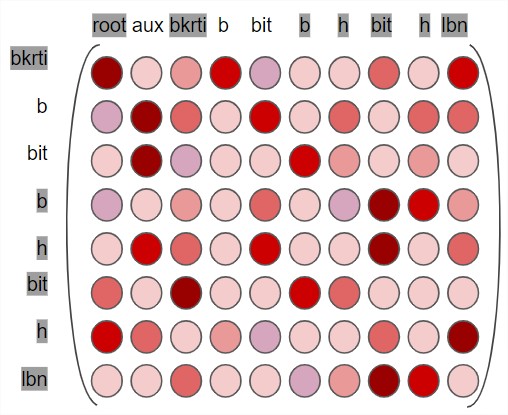}
    \caption{The Head matrix of the Hebrew sentence \textit{bkrti bbit hlbn}. Each row depicts the scores assigned to all heads of a particular segment (including the root and auxiliary tokens). The darker color indicates a higher score. The input to \citeauthor{dozatmanning}'s original architecture consists of the root and gray-marked segments.}
    \label{fig:matrix}
\end{figure}

\paragraph{The Joint Arc-Factored Model}
The central component of the architecture is an arc-factored model capable of selecting the highest-scoring subset from the (constrained) set of arcs. 
Our departure point is the Biaffine-Score architecture of \citet{dozatmanning} , which is in turn based on \citet{kiperwassergoldberg}, and is currently the de factor standard architecture for dependency parsing. 
In order to turn this architecture into a joint segmentation-parsing  model, we introduce several novelties into   \citet{dozatmanning}. 

In the original architecture, the input consists of a tokenized sentence with an additional root token. However, for joint prediction, we modify the input to be the linearized lattice of the input sentence and add two nodes: the {\em root node} and an {\em auxiliary node}. 
The linearized lattice 
represents the list of all segments from all possible analyses, ordered as detailed in Section \ref{linerization}. 

During training, the head of any segment that does not appear in the gold dependency tree is attached under the {\em auxiliary} token, and thus, the model learns to assign to the {\em root} only the segments of the relevant morphological analyses in context. At inference times, the irrelevant segments are assigned to the auxiliary token, and the auxiliary subtree is removed altogether, leaving a single rooted dependency tree intact.
Figure \ref{fig:matrix} describes the head matrix of the Hebrew sentence '\textit{bkrti bbit hlbn}' in our model. The gray-marked segments participate in the final tree. 

\paragraph{Input Embeddings} \label{embedding:para}
The input to our proposed architecture consists of the contextualized embeddings of the segments sequence, including also the root and auxiliary tokens at the beginning of that sequence.
Unlike the original input, the linearized lattice representation lacks a coherent context for generating high-quality contextualized embeddings. Therefore, we establish a valid context for each token analysis from the original sentence, and we employ contextualized embedding that reflect this context.

To create the embeddings for each of the input segments, we begin with the original sequence of tokens  $s_1, s_2, \ldots, s_k$. For each analysis $a_{j}^{i}$, we create an analysis where we replace $s_{j}$ with the sequence of morphological segments 
and results in the following sequence:
\begin{gather*}
s_1,\dots,s_{j-1}, m_{(i,j)}^{1},\dots,m_{(i,j)}^{r_{(i,j)}},s_{j+1},\dots,s_k
\end{gather*}
Using this modified context, we obtain contextual embeddings for each of $m_{(i,j)}^{1}, m_{(i,j)}^{2}, \ldots, m_{(i,j)}^{r_{(i,j)}}$ using an LLM encoder. 
We apply the LLM's original tokenizer to the morpheme sequence. if the morpheme is present in the LLM's encoder vocabulary, it remains untokenized and receives a single vector embedding. Conversely, for out-of-vocabulary morpheme, tokenization is carried out based on the LLM's encoder. Each new token receives a vector, and the vector of the first token represents the whole original morpheme.

The embeddings of the root and auxiliary tokens are directly derived from the original token sequence.\footnote{During the embedding process, we may generate different embedding vectors for segments with identical forms, e.g., the morph \(b\) repeats twice in the matrix in Figure \ref{fig:matrix}. However, these identical  forms reside in the context of different token analyses, and thus their contextualized embeddings are different.}

\paragraph{The Arc Selection Phase}
The encoded segments are inputted into \citeauthor{dozatmanning}'s architecture, which produces two matrices: one for head prediction and another for label prediction. The head prediction matrix assigns a probability to each pair of segments, including the root and auxiliary tokens, indicating whether one is the head of the other. A similar process is carried out for each pair of tokens with respect to every possible label.
The architecture by \citeauthor{dozatmanning} (\citeyear{dozatmanning}) for dependency parsing remains unchanged, and the matrices are employed to represent the syntactic relationships between segments. The introduction of an auxiliary token allows for the exclusion of specific segments from the final tree by removing all segments for which it serves as the head. All other segments are retained as nodes in the final tree. This modification enables the architecture to perform joint segmentation and parsing predictions.

\paragraph{Input Constraints}
To ensure that the output conforms to the constraints outlined in section \ref{para:joint_prediction}, it is imperative to limit the segment-sets that can form trees constructed beneath the root.

To implement the {\em constrained} function, we adopt a strategy where only one analysis per token is selected in each possible tree. In cases where the highest scoring subset selects more than one analysis per token, or if no segments from its analyses were selected, we opt for the next best analysis, which contains the head segment  with the highest score. When a particular analysis is chosen, we mask the auxiliary token for each of the segments, ensuring that all of them are included in the final tree. Finally, we mask all arcs where  a segment in a chosen analysis that relates to a segment in a different analysis of the same token.

\paragraph{Multitask Learning}
We aimed to leverage additional linguistic tasks such as gender, person, number, and POS. Consequently, we expanded upon the original architecture introduced by \citeauthor{dozatmanning} to accommodate these MTL objectives. The input embedding is processed through a BiLSTM, and the output is utilized by both the aforementioned joint architecture and the MTL architecture designed to handle these linguistic tasks.

\begin{figure}[t]
    \centering
    \includegraphics[width=0.45\textwidth]{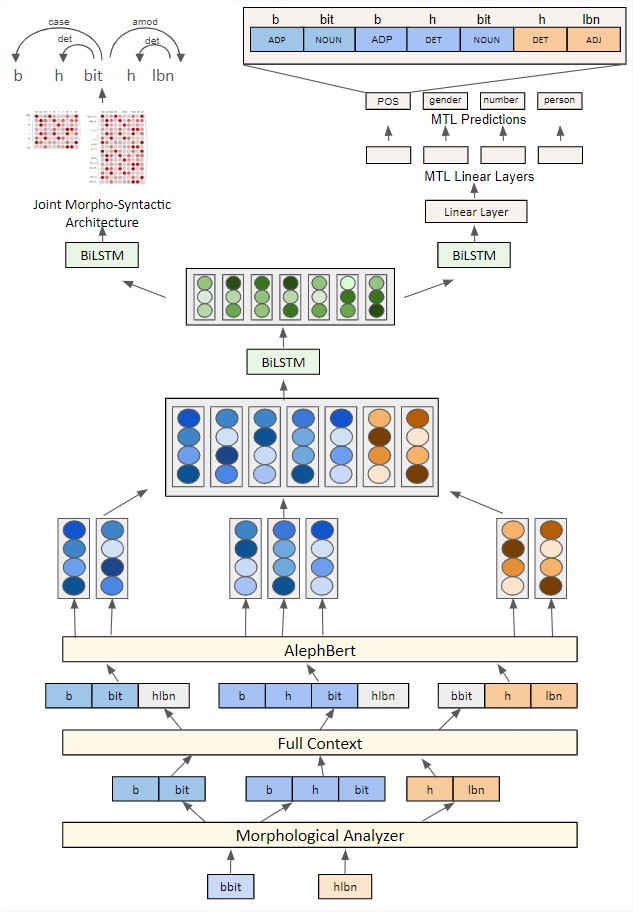}
    \caption{The comprehensive architecture examines the phrase '\textit{bbit hlbn}' (in-the-house the-white), encompassing the processes of morphological analysis, generating context for each analysis, acquiring contextualized embeddings, constructing a dependency tree, and predicting linguistic features.}
    \label{fig:architecture}
\end{figure}

\paragraph{The Overall Architecture}  
Figure \ref{fig:architecture} presents the  proposed architecture from a bird's eye view. We illustrate it for the phrase '\textit{bbit hlbn}' ("in the white house") from Figure \ref{fig:matrix} which presented the head matrix of the phrase.
The architecture of our model begins by taking a sentence and embedding its linearized lattice representation. These embeddings then undergo processing through a two-layer BiLSTM. The BiLSTM's output is further directed into two distinct BiLSTMs: one for the Biaffine score architecture and another for the MTL architecture. Within the Biaffine module, it is utilized to generate head and label matrices, facilitating the prediction of a well-structured dependency tree. In the MTL segment, the output undergoes dimension reduction through a linear layer. Subsequently, an additional and separate linear layer is applied for each MTL task (POS, gender, number, and person) to predict language features. We compute the loss using the \textit{cross-entropy} function for the head, label, and each of the MTL tasks, and then aggregate them into a combined loss.

\section{Experimental Setup}
\paragraph{Goal}
We set out to evaluate the performance of the proposed joint architecture on segmentation, tagging and parsing. In all experiments, we show the segmentation (SEG) and dependency parsing (DEP) $F_1 score$. Additionally, for experiments with the POS MTL, we present the POS $F_1 score$.

\paragraph{Data}
All our experiments were trained and tested on the standard split of the Hebrew section of the UD treebank collection \cite{universaldependencies}. The training set, dev set, and test set  consist of 5,168 sentences, 484 sentences, and 491 sentences, respectively. The morphological analysis for the input of our model is provided by the Morphological Analyzer (MA) of \citet{yap}.\footnote{\url{https://github.com/OnlpLab/yap}} The MA provides the segmentation, Part-of-Speech (POS) tags and morphological features for each segment in each one of the possible analyses.\footnote{To address situations where a segment has multiple potential POS tags or morphological features, we employ a  criterion based on  the most common label (or the first one in case of a tie).}

\paragraph{Embeddings}
The way we generate embedding  for the input lattice complements the architectural design and significantly impacts the parser performance. Alongside our proposed sentence-contextualized embedding (Section \ref{embedding:para}), we assessed two alternative techniques. We tested this structure with AlephBert's \cite{alephbert} static (Static) enbeddings and contextualized (Contextualized) embeddings generated directly for the linearized lattices. 

\paragraph{Evaluation Scenarios}
As part of our model we use a Morphological Analyzer (MA) component for generating the lattices. However, any realistic MA is not exhaustive, as it may lack some analyses, with certain tokens  entirely missing (out of vocabulary, OOV). Hence, we aim to gauge the effect of the MA coverage on the parser performance.
In the \textbf{Infused} scenario, we experiment in a setup where for all sentences the correct analysis is guaranteed to be incorporated as one of the lattice's internal paths. To establish the infused scenario, we examine all tokens in the dataset and integrate the gold analysis for each token back into the MA. In the \textbf{Uninfused} scenario, which represents a realistic scenario, we use the MA as is, and there may be missing  analyses in the constructed lattices at inference time.

\paragraph{Models}
Current SOTA results in Hebrew dependency parsing are presented by Trankit \cite{trankit} and Stanza \cite{stanza}, both of which are multilingual neural parsers. Since our proposed architecture essentially extends that of \citet{dozatmanning}, we also evaluate their architecture in a pure pipeline setting.  In this paper, we  introduce three  variations of Hebrew parsing  employing distinct segmentation techniques, as described below.
\begin{itemize}
    \item[] \(\circ\) \textbf{Gold:} The pipeline accepts the gold segmentation from the Hebrew  treebank.
    \\ \(\circ\) \textbf{Predicted:} The pipeline accepts the SOTA segmentation predicted by \citet{segmentor}'s segmention model.
    \\ \(\circ\) \textbf{Joint:} The joint scenario  infers both the segmentation and the parse tree using the proposed architecture.
\end{itemize}
For each baseline we present SEG, POS and dependency parsing DEP scores. 


\paragraph{Metrics}
The Labeled Attachment Score (LAS) serves as the predominant metric for measuring dependency parsing accuracy. 
However, this measurement method is incompatible for the complex segmentation task associated with Morphologically Rich Languages (MRLs) since the predicted segments (i.e., nodes in the tree) may differ from the gold ones.
For this reason,  we evaluate segmentation  using the \textit{aligned multi-set} $F_1 score$ (\citet{segmentor}, \citet{idansegmentor}) metric, specifically chosen for to cope with cases where gold and predicted segmentations do not align, and also caters for backwards compatibility with previous work. 
All results we present are  averaged over five distinct experiments with random seeds.

\section{Results and Analysis}
\begin{table}[t]
\centering
\scalebox{0.9}{
\begin{tabular}{l|ll}
\hline
 & \textbf{SEG} & \textbf{DEP}\\
\hline
 Biaffine + Oracle SEG & 100 & 86.76\\
 Biaffine + Predicted SEG & 97.6 & 71.57\\
\hline
\end{tabular}}
\caption{\label{table:oracle}
The original Biaffine architecture of \citet{dozatmanning} with gold and predicted SEG.
}
\end{table}

\begin{table}[t]
\centering
\scalebox{0.8}{
\begin{tabular}{l|lll}
\hline
Model \(\rightarrow\) & \multicolumn{3}{c}{\textbf{Stanza/Trankit}}\\
Input \(\downarrow\) & SEG & POS & DEP\\
\hline
\hline
 {Oracle SEG} & 100/100 & 94.75/97.2 & 78.38/89.42\\
 Model SEG & 89.51/95.2 & 85.03/92.68 & 67.45/{83.55}\\
 Predicted SEG & 97.6/97.6 & 92.73/94.92 & 75.52/\textbf{85.66}\\
\hline
\end{tabular}}
\caption{\label{table:baseline}
Trankit \cite{trankit} and Stanza \cite{stanza} results for SEG, POS and DEP parsing in Hebrew. Oracle provides gold segments, Model provides the internal segmentation of Stanza/Trankit, and Predicted is the SOTA segmentation of AlephBERT \cite{alephbert}.
}
\end{table}

Table \ref{table:oracle} demonstrates the performance of the \citet{dozatmanning} architecture using both gold \textit{oracle} and predicted segmentation as input to the biaffine architecture. 
These results establish that when not using the gold ({\em oracle}) segmentation, even a small drop in segmentation leads to a substantial decline in dependency parsing accuracy, thereby emphasizing the importance of segmentation in parsing.

\begin{table}[t]
\centering
\scalebox{0.9}{
\begin{tabular}{l|lll}
\hline
 & \textbf{SEG} & \textbf{POS} & \textbf{DEP}\\
\hline
 No MTL & 97.68 & - & 84.69\\
 + Gender & 97.67 & - & 84.88\\
 + Person & 97.61 & - & 84.99\\
 + Number & 97.75 & - & 84.76\\
 + POS & \textbf{97.71} & 94.41 & \textbf{85.45}\\
 + POS (heBERT)  & 97.51 & 93.9 & 84.31\\
 + POS (mBERT) & 96.84 & 91.8 & 80.68 \\
 All MTLs & 97.68 & 94.31 & 85\\
\hline
\end{tabular}}
\caption{\label{table:all}
Model evaluation under three conditions: without employing any MTL, utilizing one MTL at a time, and incorporating all MTL components simultaneously.
}
\end{table}

\begin{table}[t]
\centering
\scalebox{0.9}{
\begin{tabular}{l|ll}
\hline
 & \textbf{SEG} & \textbf{DEP}\\
\hline
 Static & 96.36 & 78.19\\
 Contextualized & 97.18 & 82.23\\
 Proposed & 97.68 & 84.69\\
\hline
\end{tabular}}
\caption{\label{table:embeddings}
Our model incorporates various embedding representations. The `Static' and `Contextualized' embeddings use a lattice context,  `Proposed'  uses a valid sentence context for each analysis. 
}
\end{table}

Table \ref{table:baseline}   shows the results of Trankit \cite{trankit} and Stanza \cite{stanza} compared with our proposed model.\footnote{Our models and code are publicly available at \url{https://github.com/OnlpLab/Hebrew-Dependency-Parsing}. All hyperparameters are listed in the Appendix.}  Prior to this work, Trankit achieved state-of-the-art results on Hebrew parsing. The Oracle segmentation scenarios of Trankit and Stanza provide an idealized and unrealistic scenario, with a substantial drop when moving to non-gold scenarios. Notably, the experimental results of Trankit with our suggested external Hebrew segmentation sets a new SOTA to which our architecture achieves comparable results. The difference is minor, yet our proposed architecture stands out by offering an efficient full pipeline that delivers segmentation, tagging and parsing simultaneously, avoiding the need to train, maintain, and install modules separately. 


Table \ref{table:all} shows the results of our proposed approach with ablation of the MTL contribution. These results demonstrate that our joint architecture surpasses the original Biaffine architecture in Hebrew parsing, attaining a state-of-the-art (SOTA) performance with an 85.45 $F_1 score$, better than the parsing frameworks of Stanza and Trankit. The results are comparable for the combination of Trankit with an external model with a separately trained decoder, with different training regimes, while in our model, SEG, POS and DEP  are trained, and predicted, jointly.
Furthermore, Table \ref{table:embeddings} highlights the significance of the embedding method used for encoding the input lattices. While a substantial improvement is evident between static and contextualized embeddings, a notable enhancement is also observed when altering the context of the linearized lattice as we propose. 

Table \ref{table:infused}  illustrates the extent to which limitations of the MA component affect parsing performance, in cases where certain analyses may be absent for some tokens at inference time. 
It is evident that when the correct analyses are included in the set of possible analyses, it selects a better segmentation that results in more accurate parsing. So, improvement of the MA coverage is expected to yield even further improvement in parsing.

Finally, since the LLM may be seamlessly replaced,  further improvement may come from a better LLM encoder. Table \ref{table:all} shows that replacing mBERT \cite{mbert} with the AlephBERT \cite{alephbert} encoder gave a significantly improved performance. This leaves a promise of further improving performance with significantly better   LLMs.

\begin{table}[t]
\centering
\scalebox{0.9}{
\begin{tabular}{l|lll}
\hline
 & \textbf{SEG} & \textbf{POS} & \textbf{DEP}\\
 \hline
 Infused & 98.47 & - & 85.56\\
 Infused  + MTL POS  & 98.52 & 95.22 & \textbf{86.55}\\
\hline
 Uninfused & 97.68 & - & 84.69 \\
 Uinfused + MTL POS & 97.71 & 94.41 & 85.45\\
 \hline
 \end{tabular}}
\caption{\label{table:infused}
Our proposed model with infused MA, with and without POS MTL. 
}
\end{table}

\paragraph{Error Analysis} We performed a manual error analysis on a subset of 50 sentences from the Hebrew UD HTB dev set. In these, there are merely 8 segmentation errors, with 5 of being a missing definite article ({\em 'he hydia'h'}) and the remaining 3 involving incorrect segmentation of fused suffixes. In addition, a total of 108 dependency errors were identified, classified into four  categories: prediction errors, wrong gold, truly ambiguous, and others (Table \ref{table:error types} in the Appendix). Of these, 70\% are prediction errors.
We categorized the errors based on the dependency labels that are involved. The predominant error type is associated with PP attachment, where 20\% of the errors confuse the {\em obl} and {\em nmod} relations, indicating a confusion between the complements of the verb and modifiers of the noun, respectively (see further details in Table \ref{table:label errors} in the Appendix).

\section{Related and Future Work}
Previous research has delved into lattice-based dependency parsing for MRLs such as Hebrew \cite{yap}, Turkish \cite{turkish}, and Sanskrit \cite{krishna-etal-2020-graph}. However, these prior contributions predominantly utilized graph-based and transition-based systems grounded in feature functions that are hand-engineered. In contrast, our current work takes a different perspective, presenting a purely neural architecture. A distinct challenge lies in generating embeddings for the lattice arcs, which represent a non-linear structure --- an atypical input signal for language models. 
The aforementioned lattice-based parsing architectures do not attend to this complexity 
thereby missing out on the advantages offered by contemporary Large Language Models (LLMs). This paper bypasses this divide, proposing an approach that effectively handles the intricate context and creates robust representations for lattices using neural encoders.

While neural studies in MRL parsing, such as the work by \citet{sandhan2021little}, also leverage the Biaffine architecture of \citet{dozatmanning}, they typically focus on architectures that handles segmented and unambiguous inputs. Consequently, these models do not cope well with the challenges posed by the vast number of ambiguous words prevalent in MRLs such as Hebrew. In contrast, our proposed architecture accommodates ambiguous input, offering a unified solution that addresses both segmentation, morphological disambiguation, and parsing, in a single model.

In future research, we aim to assess our proposed framework on other MRLs, evaluate its performance across various language types and assess its generalization capabilities for lower-resourced languages. Additionally, we aim to explore further enhancements of MTL in parsing, by adding joint semantic predictions such as NER and SRL.
\section{Conclusions}
In this paper we present a novel neural framework for jointly segmenting and  parsing morpho-syntactic structures in Morphologically Rich Languages (MRLs). We address the intricate and complex nature of words in these languages and propose a method for incorporating linguistic information structured in a lattice into a neural parsing architecture. The contribution of this paper is  manifold. First, we provide a language-agnostic neural joint architecture that can be used to confirm or disprove the joint hypothesis juxtaposed in the pre-neural era for MRLs. Second, we provide a thorough empirical investigation of Hebrew, providing SOTA results using a single joint model. Finally, as the proposed architecture relies on an LLM encoder, advances are expected to be achieved as LLMs  further improve  for low- and medium-resources MRLs, potentially closing the gap with non-MRLs.
\section{Limitations}
In our study, we introduce a joint morpho-syntactic architecture tailored to address the segmentation and parsing challenges of Morphologically Rich Languages (MRLs) in a single a model.  It is important to note that the term ``segmentation" can have various meanings, and in our work, we specifically refer to the segmentation of raw tokens into multiple  meaning-bearing units, each of which carrying its own POS tag. This is compatible with previous work on Hebrew and other Semitic languages \cite{adler-elhadad-2006-unsupervised,alephbert}.
All  modeling and design decisions made are language-agnostic.  Having acknowledged that, we conducted experiments using Hebrew as our test language. This investigation can and should be extended to any language that has a UD treebank and a wide-coverage morphological analyzer (MA). 

One of the key components of our approach is the Morphological Analyzer (MA), which provides a list of possible analyses for each token. This component is not always freely  available. Here, our experiments focused on Hebrew. It is noteworthy however that MAs are available for many languages and specifically for MRLs \cite{conll-ul}. MAs are available also for Arabic \cite{arabic-ma}, Turkish \cite{turkish-ma} and Sanskrit.\footnote{\url{http://sanskrit.jnu.ac.in/morph/analyze.jsp}} 
It is also worth noting that the open MAs we can access is academia are fairly small, but there exist larger lexical MAs in the industry, for Hebrew and  other languages.\footnote{For instance  through   \url{https://lexicala.com/}}
On top of that, creating proper contextualized embeddings for each segment in the lattice is more time-consuming than is desired, and in  future work we aim to specifically address these efficiency concerns.

Finally, when generating contextualized embeddings for the input lattice we employed AlephBert, a  pre-trained monolingual language model for Hebrew.  Substituting this model with a bigger or more advanced model can potentially yield further improvements. 
More work may be done on improving the way we encode the linearized lattices, either in the realm of pre-tuning, or by fine-tuning the LLM specifically for the lattice-encoding task.

\section*{Acknowledgements}
We thank Eylon Gueta, Refael Shaked Greenfeld and Amit Seker for fruitful discussions. We also  thank the audience of the NLP-BIU seminar and three anonymous reviewers for thoughtful comments on earlier drafts. This research was funded by the European Research Council (ERC) grant number 677352 and a Ministry of science and education (MOST) grant number 3-17992, for which we are grateful. The research was further supported by a grant from the Israeli Innovation Authority (KAMIN), and  computing resources kindly funded by a VATAT grant and the Data Science Institute at Bar-Ilan University (BIU-DSI).

\begin{singlespace}
\bibliography{biblio}

\begin{thebibliography}{30}
\expandafter\ifx\csname natexlab\endcsname\relax\def\natexlab#1{#1}\fi

\bibitem[{Adler and Elhadad(2006)}]{adler-elhadad-2006-unsupervised}
Meni Adler and Michael Elhadad. 2006.
\newblock \href {https://doi.org/10.3115/1220175.1220259} {An unsupervised morpheme-based {HMM} for {H}ebrew morphological disambiguation}.
\newblock In \emph{Proceedings of the 21st International Conference on Computational Linguistics and 44th Annual Meeting of the Association for Computational Linguistics}, pages 665--672, Sydney, Australia. Association for Computational Linguistics.

\bibitem[{Brusilovsky and Tsarfaty(2022)}]{idansegmentor}
Idan Brusilovsky and Reut Tsarfaty. 2022.
\newblock Neural token segmentation for high token-internal complexity.
\newblock In \emph{arXiv:2203.10845v1}.

\bibitem[{Cohen and Smith(2007)}]{cohen-smith-2007}
Shay~B. Cohen and Noah~A. Smith. 2007.
\newblock \href {https://aclanthology.org/D07-1022} {Joint morphological and syntactic disambiguation}.
\newblock In \emph{Proceedings of the 2007 Joint Conference on Empirical Methods in Natural Language Processing and Computational Natural Language Learning ({EMNLP}-{C}o{NLL})}, pages 208--217, Prague, Czech Republic. Association for Computational Linguistics.

\bibitem[{Dozat and Manning(2017)}]{dozatmanning}
Timothy Dozat and Christopher~D. Manning. 2017.
\newblock Deep biaffine attention for neural dependency parsing.
\newblock In \emph{In International Conference on Learning Representations (ICLR)}.

\bibitem[{Galley and Manning(2009)}]{machine-translation}
Michel Galley and Christopher~D. Manning. 2009.
\newblock \href {https://aclanthology.org/P09-1087} {Quadratic-time dependency parsing for machine translation}.
\newblock In \emph{Proceedings of the Joint Conference of the 47th Annual Meeting of the {ACL} and the 4th International Joint Conference on Natural Language Processing of the {AFNLP}}, pages 773--781, Suntec, Singapore. Association for Computational Linguistics.

\bibitem[{Garimella et~al.(2021)Garimella, Chiticariu, and Li}]{question-answering}
Aparna Garimella, Laura Chiticariu, and Yunyao Li. 2021.
\newblock Domain-aware dependency parsing for questions.
\newblock pages 4562--4568.

\bibitem[{Goldberg and Tsarfaty(2008)}]{goldberg-tsarfaty-2008}
Yoav Goldberg and Reut Tsarfaty. 2008.
\newblock \href {https://aclanthology.org/P08-1043} {A single generative model for joint morphological segmentation and syntactic parsing}.
\newblock In \emph{Proceedings of ACL-08: HLT}, pages 371--379, Columbus, Ohio. Association for Computational Linguistics.

\bibitem[{Green and Manning(2010)}]{green-manning-2010}
Spence Green and Christopher~D. Manning. 2010.
\newblock \href {https://aclanthology.org/C10-1045} {Better {A}rabic parsing: Baselines, evaluations, and analysis}.
\newblock In \emph{Proceedings of the 23rd International Conference on Computational Linguistics (Coling 2010)}, pages 394--402, Beijing, China. Coling 2010 Organizing Committee.

\bibitem[{Honnibal and Montani(2017)}]{spacy}
Matthew Honnibal and Ines Montani. 2017.
\newblock {spaCy 2}: Natural language understanding with {B}loom embeddings, convolutional neural networks and incremental parsing.
\newblock To appear.

\bibitem[{Hwang et~al.(2020)Hwang, Yim, Park, Yang, and Seo}]{information-extraction}
Wonseok Hwang, Jinyeong Yim, Seunghyun Park, Sohee Yang, and Minjoon Seo. 2020.
\newblock \href {http://arxiv.org/abs/2005.00642} {Spatial dependency parsing for 2d document understanding}.
\newblock \emph{CoRR}, abs/2005.00642.

\bibitem[{Kiperwasser and Goldberg(2016)}]{kiperwassergoldberg}
Eliyahu Kiperwasser and Yoav Goldberg. 2016.
\newblock Simple and accurate dependency parsing using bidirectional lstm feature representations.
\newblock In \emph{Transactions of the Association for Computational Linguistics}, volume~4, pages 313--327.

\bibitem[{Kondratyuk and Straka(2019{\natexlab{a}})}]{udpipe}
Dan Kondratyuk and Milan Straka. 2019{\natexlab{a}}.
\newblock 75 languages, 1 model: Parsing universal dependencies universally.
\newblock In \emph{In Proceedings of the 2019 Conference on Empirical Methods in Natural Language Processing and the 9th International Joint Conference on Natural Language Processing (EMNLPIJCNLP)}, pages 2779--2795.

\bibitem[{Kondratyuk and Straka(2019{\natexlab{b}})}]{udify}
Dan Kondratyuk and Milan Straka. 2019{\natexlab{b}}.
\newblock \href {https://aclanthology.org/D19-1279} {75 languages, 1 model: Parsing {U}niversal {D}ependencies universally}.
\newblock In \emph{Proceedings of the 2019 Conference on Empirical Methods in Natural Language Processing and the 9th International Joint Conference on Natural Language Processing (EMNLP-IJCNLP)}, pages 2779--2795, Hong Kong, China. Association for Computational Linguistics.

\bibitem[{Krishna et~al.(2020{\natexlab{a}})Krishna, Gupta, Garasangi, Satuluri, and Goyal}]{krishna-etal-2020-keep}
Amrith Krishna, Ashim Gupta, Deepak Garasangi, Pavankumar Satuluri, and Pawan Goyal. 2020{\natexlab{a}}.
\newblock \href {https://doi.org/10.18653/v1/2020.emnlp-main.388} {Keep it surprisingly simple: A simple first order graph based parsing model for joint morphosyntactic parsing in {S}anskrit}.
\newblock In \emph{Proceedings of the 2020 Conference on Empirical Methods in Natural Language Processing (EMNLP)}, pages 4791--4797, Online. Association for Computational Linguistics.

\bibitem[{Krishna et~al.(2020{\natexlab{b}})Krishna, Santra, Gupta, Satuluri, and Goyal}]{krishna-etal-2020-graph}
Amrith Krishna, Bishal Santra, Ashim Gupta, Pavankumar Satuluri, and Pawan Goyal. 2020{\natexlab{b}}.
\newblock \href {https://doi.org/10.1162/coli_a_00390} {A graph-based framework for structured prediction tasks in {S}anskrit}.
\newblock \emph{Computational Linguistics}, 46(4):785--845.

\bibitem[{Libovick{\'{y}} et~al.(2019)Libovick{\'{y}}, Rosa, and Fraser}]{mbert}
Jindrich Libovick{\'{y}}, Rudolf Rosa, and Alexander Fraser. 2019.
\newblock \href {http://arxiv.org/abs/1911.03310} {How language-neutral is multilingual bert?}
\newblock volume abs/1911.03310.

\bibitem[{Minh~Nguyen and Nguyen(2021)}]{trankit}
Amir Pouran Ben~Veyseh Minh~Nguyen, Viet~Lai and Thien~Huu Nguyen. 2021.
\newblock Trankit: A light-weight transformer-based toolkit for multilingual natural language processing.
\newblock In \emph{In Proceedings of the 16th Conference of the European Chapter of the Association for Computational Linguistics: System Demonstrations}, volume 322, pages 80--90.

\bibitem[{More et~al.(2019)More, Seker, Basmova, and Tsarfaty}]{yap}
Amir More, Amit Seker, Victoria Basmova, and Reut Tsarfaty. 2019.
\newblock \href {https://aclanthology.org/Q19-1003} {Joint transition-based models for morpho-syntactic parsing: Parsing strategies for {MRL}s and a case study from {M}odern {H}ebrew}.
\newblock volume~7, pages 33--48, Cambridge, MA. MIT Press.

\bibitem[{More et~al.(2018)More, Çetinoğlu, Çagri Ç{\"o}ltekin, Habash, Sagot, Seddah, Taji, and Tsarfaty}]{conll-ul}
Amir More, {\"O}zlem Çetinoğlu, Çagri Ç{\"o}ltekin, Nizar Habash, Beno{\^i}t Sagot, Djam{\'e} Seddah, Dima Taji, and Reut Tsarfaty. 2018.
\newblock \href {https://api.semanticscholar.org/CorpusID:21721927} {Conll-ul: Universal morphological lattices for universal dependency parsing}.
\newblock In \emph{International Conference on Language Resources and Evaluation}.

\bibitem[{Nivre et~al.(2016)Nivre, de~Marneffe, Ginter, Goldberg, Haji{\v{c}}, Manning, McDonald, Petrov, Pyysalo, Silveira, Tsarfaty, and Zeman}]{universaldependencies}
Joakim Nivre, Marie-Catherine de~Marneffe, Filip Ginter, Yoav Goldberg, Jan Haji{\v{c}}, Christopher~D. Manning, Ryan McDonald, Slav Petrov, Sampo Pyysalo, Natalia Silveira, Reut Tsarfaty, and Daniel Zeman. 2016.
\newblock \href {https://aclanthology.org/L16-1262} {{U}niversal {D}ependencies v1: A multilingual treebank collection}.
\newblock In \emph{Proceedings of the Tenth International Conference on Language Resources and Evaluation ({LREC}'16)}, pages 1659--1666, Portoro{\v{z}}, Slovenia. European Language Resources Association (ELRA).

\bibitem[{Qi et~al.(2020)Qi, Zhang, Zhang, Bolton, and Manning}]{stanza}
Peng Qi, Yuhao Zhang, Yuhui Zhang, Jason Bolton, and Christopher~D. Manning. 2020.
\newblock Stanza: A python natural language processing toolkit for many human languages.
\newblock In \emph{In Proceedings of the 58th Annual Meeting of the Association for Computational Linguistics: System Demonstrations}, pages 101--108.

\bibitem[{Sandhan et~al.(2021)Sandhan, Krishna, Gupta, Behera, and Goyal}]{sandhan2021little}
Jivnesh Sandhan, Amrith Krishna, Ashim Gupta, Laxmidhar Behera, and Pawan Goyal. 2021.
\newblock A little pretraining goes a long way: A case study on dependency parsing task for low-resource morphologically rich languages.
\newblock \emph{arXiv preprint arXiv:2102.06551}.

\bibitem[{Seeker and {\c{C}}etino{\u{g}}lu(2015)}]{turkish}
Wolfgang Seeker and {\"O}zlem {\c{C}}etino{\u{g}}lu. 2015.
\newblock \href {https://aclanthology.org/Q15-1026} {A graph-based lattice dependency parser for joint morphological segmentation and syntactic analysis}.
\newblock volume~3, pages 359--373, Cambridge, MA. MIT Press.

\bibitem[{Seker and Tsarfaty(2020)}]{segmentor}
Amit Seker and Reut Tsarfaty. 2020.
\newblock \href {https://aclanthology.org/2020.findings-emnlp.391} {A pointer network architecture for joint morphological segmentation and tagging}.
\newblock In \emph{Findings of the Association for Computational Linguistics: EMNLP 2020}, pages 4368--4378, Online. Association for Computational Linguistics.

\bibitem[{Seker and Tsarfaty(2021)}]{alephbert}
Amit Seker and Reut Tsarfaty. 2021.
\newblock Alephbert: A hebrew large pretrained language model to start-off your hebrew nlp application with.
\newblock In \emph{arXiv:2104.04052}.

\bibitem[{Taji et~al.(2018)Taji, Khalifa, Obeid, Eryani, and Habash}]{arabic-ma}
Dima Taji, Salam Khalifa, Ossama Obeid, Fadhl Eryani, and Nizar Habash. 2018.
\newblock \href {https://doi.org/10.18653/v1/W18-5816} {An {A}rabic morphological analyzer and generator with copious features}.
\newblock In \emph{Proceedings of the Fifteenth Workshop on Computational Research in Phonetics, Phonology, and Morphology}, pages 140--150, Brussels, Belgium. Association for Computational Linguistics.

\bibitem[{Tsarfaty(2006)}]{joint_hypothesis}
Reut Tsarfaty. 2006.
\newblock \href {https://aclanthology.org/P06-3009} {Integrated morphological and syntactic disambiguation for {M}odern {H}ebrew}.
\newblock In \emph{Proceedings of the {COLING}/{ACL} 2006 Student Research Workshop}, pages 49--54, Sydney, Australia. Association for Computational Linguistics.

\bibitem[{Tsarfaty et~al.(2020)Tsarfaty, Bareket, Klein, and Seker}]{tsarfaty-etal-2020-spmrl}
Reut Tsarfaty, Dan Bareket, Stav Klein, and Amit Seker. 2020.
\newblock \href {https://doi.org/10.18653/v1/2020.acl-main.660} {From {SPMRL} to {NMRL}: What did we learn (and unlearn) in a decade of parsing morphologically-rich languages ({MRL}s)?}
\newblock In \emph{Proceedings of the 58th Annual Meeting of the Association for Computational Linguistics}, pages 7396--7408, Online. Association for Computational Linguistics.

\bibitem[{Tsarfaty et~al.(2010)Tsarfaty, Seddah, Goldberg, Kuebler, Versley, Candito, Foster, Rehbein, and Tounsi}]{tsarfaty-etal-2010-statistical}
Reut Tsarfaty, Djam{\'e} Seddah, Yoav Goldberg, Sandra Kuebler, Yannick Versley, Marie Candito, Jennifer Foster, Ines Rehbein, and Lamia Tounsi. 2010.
\newblock \href {https://aclanthology.org/W10-1401} {Statistical parsing of morphologically rich languages ({SPMRL}) what, how and whither}.
\newblock In \emph{Proceedings of the {NAACL} {HLT} 2010 First Workshop on Statistical Parsing of Morphologically-Rich Languages}, pages 1--12, Los Angeles, CA, USA. Association for Computational Linguistics.

\bibitem[{Y{\i}ld{\i}z et~al.(2019)Y{\i}ld{\i}z, Avar, and Ercan}]{turkish-ma}
Olcay~Taner Y{\i}ld{\i}z, Beg{\"u}m Avar, and G{\"o}khan Ercan. 2019.
\newblock \href {https://doi.org/10.26615/978-954-452-056-4_156} {An open, extendible, and fast {T}urkish morphological analyzer}.
\newblock In \emph{Proceedings of the International Conference on Recent Advances in Natural Language Processing (RANLP 2019)}, pages 1364--1372, Varna, Bulgaria. INCOMA Ltd.

\end{thebibliography}
\end{singlespace}
\appendix

\section{Appendix}
\subsection{Hyperparameters and Computing Resources}
For all models we used the hyper parameters in Table~\ref{table:hyper}. The research was conducted using a {\em NVIDIA GeForce GTX 1080 Ti} machine. To enhance time efficiency, we pre-generated embeddings before the training phase, avoiding the need to create them for each epoch. The process of generating embeddings for the entire training dataset took approximately 80 minutes. On average, each epoch lasted 15 seconds, resulting in a total training time of approximately 7 minutes.

For evaluation purposes, we assessed the efficiency of both embedding and inference on the test dataset, where the longest sentence consisted of 61 tokens with linearized lattice of 217 morphemes and the shortest contained 2 tokens with linearized lattice of 4 morphemes. The average linearized lattice contains 57 morphemes. The average time for embedding was 0.24 seconds, and for inference, it was 0.017 seconds. The maximum time recorded for embedding was 0.94 seconds, and for inference, it was 0.19 seconds. We acknowledge the efficiency bottleneck at the embedding generation phase, which we reserve for future research. 

\begin{table}[t]
\centering
\scalebox{0.8}{
\begin{tabular}{l|l}
\hline
 Embedding dimension & 768\\
 BiLSTM hidden size & 600\\
 Batch size & 32\\
 Embedding dropout & 0.3\\
 ARC MLP dropout & 0.3\\
 Label MLP dropout & 0.3\\
 All BiLSTMs depth & 1\\
 MLP depth & 1\\
 Arc MLP size & 500\\
 Label MLP size & 100\\
 Learning rate & 0.001\\
 MTL linear layer size & 600\\
 
\hline
\end{tabular}}
\caption{\label{table:hyper}
Hyperparameter Settings
}
\end{table}

\subsection{Error Analysis}
We performed a manual error analysis by an expert on 50  sentences sampled from the dev set, and found 108 parsing errors.
Table \ref{table:error types} presents the types of dependency errors, where 70\% are prediction errors and the rest are not considered parser errors by the expert.

\begin{table}[t]
\centering
\scalebox{0.8}{
\begin{tabular}{l|ll}
\hline
 & \textbf{number} & \textbf{percent}\\
\hline
 prediction error & 76 & 70\%\\
 gold error & 13 & 12\%\\
 ambiguous & 11 & 10\%\\
 other & 8 & 8\%\\
 all & 108 & 100\%\\
\hline
\end{tabular}}
\caption{\label{table:error types}
Classification of errors by type.
}
\end{table}

Table \ref{table:label errors} further presents the classification of errors by gold labels. For each label we count three types of errors: exclusively a head error, exclusively a label error, or an error encompassing both the head and label. We can see that {\em oblique} and {\em nmod} are top ranked, followed by {\em apposition, advmod} and {\em conj}.
Interestingly, at the middle of the Table we see that on top of coordination {\em conj, cc}, which is known to be challenging to disambiguate,  the construct-state construction {\em compound:smixut}, a well-known Semitic phenomenon, also appears to be confusing for the parser.

\begin{table*}[t]
\centering
\scalebox{0.9}{
\begin{tabular}{l|lllll}
\hline
\hline
\textbf{gold label} & \textbf{head} & \textbf{label} & \textbf{head + label} & \textbf{number} & \textbf{percent}\\
\hline
 obl & 4 & 1 & 6 & 11 & 10.19\%\\
 nmod & 7 & 2 & 1 & 10 & 9.26\%\\
 appos & 4 & 1 & 4 & 9 & 8.34\%\\
 advmod & 5 & 0 & 3 & 8 & 7.41\%\\
 conj & 6 & 1 & 1 & 8 & 7.41\%\\
 cc & 3 & 1 & 1 & 5 & 4.63\%\\
 ccomp & 0 & 3 & 2 & 5 & 4.63\%\\
 compound:smixut & 1 & 4 & 0 & 5 & 4.63\%\\
 dep & 0 & 2 & 3 & 5 & 4.63\%\\
 amod & 2 & 3 & 0 & 5 & 4.63\%\\
 acl:relcl & 3 & 1 & 0 & 4 & 3.7\%\\
 case & 3 & 0 & 1 & 4 & 3.7\%\\
 det & 1 & 3 & 0 & 4 & 3.7\%\\
 obj & 0 & 3 & 1 & 4 & 3.7\%\\
 nsubj & 1 & 0 & 2 & 3 & 2.78\%\\
 nmod:poss & 2 & 1 & 0 & 3 & 2.78\%\\
 fixed & 1 & 0 & 2 & 3 & 2.78\%\\
 mark & 2 & 1 & 0 & 3 & 2.78\%\\
 root & 0 & 0 & 2 & 2 & 1.9\%\\
 advcl & 1 & 1 & 0 & 2 & 1.9\%\\
 acl & 0 & 0 & 2 & 2 & 1.9\%\\
 parataxis & 0 & 1 & 0 & 1 & 0.93\%\\
 xcomp & 0 & 0 & 1 & 1 & 0.93\%\\
 flat:name & 0 & 1 & 0 & 1 & 0.93\%\\
 \hline
 Total & 46 & 30 & 32 & 108 & 100\%\\
\hline
\hline
\end{tabular}}
\caption{\label{table:label errors}
Classification of errors by gold labels. Each label is divided into three types of errors: exclusively a head error, exclusively a label error, or an error encompassing both the head and label. 
}
\end{table*}

\end{document}